\documentclass[runningheads]{llncs}
\usepackage{graphicx}
\usepackage{comment}
\usepackage{amsmath,amssymb} 
\usepackage{color}

\usepackage[width=122mm,left=12mm,paperwidth=146mm,height=193mm,top=12mm,paperheight=217mm]{geometry}

\usepackage{pgfplots}
\usepackage{tikz}

\begin{document}

\pagestyle{headings}
\mainmatter
\def\ECCVSubNumber{1225}  

\title{Back to the Future: Joint Aware Temporal Deep Learning 3D Human Pose Estimation}

\titlerunning{Back to the Future: Joing Aware Temporal Deep Learning 3D Human Pose Detection} 
\authorrunning{V. Gupta} 
\author{Vikas Gupta vikasgupta2019@u.northwestern.edu}
\institute{Northwestern University}

\titlerunning{Joint Aware Temporal Deep Learning 3D Human Pose Estimation}
%
\author{Vikas Gupta\inst{1}} 
%
\authorrunning{V. Gupta}
%
\institute{Northwestern University, Evanston, IL 60208, USA \and
\email{vikasgupta2019@u.northwestern.edu}\\
}
\maketitle

\begin{abstract}
We propose a new deep learning network that introduces a deeper CNN channel filter and constraints as losses to reduce joint position and motion errors for 3D video human body pose estimation.  
Our model outperforms the previous best result from the literature based on mean per-joint position error, velocity error, and acceleration errors on the Human 3.6M  benchmark corresponding to a new state-of-the-art mean error reduction in all protocols and motion metrics. Mean per joint error is reduced by 1\%, velocity error by 7\%  and acceleration by 13\% compared to the best results from the literature. Our contribution increasing positional accuracy and motion smoothness in video can be integrated with future end to end networks without increasing network complexity.


\end{abstract}

\section{Introduction}

In this work, we propose a joint aware deep learning network to reduce joint position and motion errors for temporal human body pose estimation. A learned representation encodes skeleton based 3D geometry information. We demonstrate a model that achieves improved 3D human pose estimation on average and over a broad set of action wise pose estimates.  
In the supervised setting, our refined joint aware fully-convolutional model outperforms the previous best result from the literature based on mean per-joint position error on the Human 3.6M benchmark, corresponding to a new state-of-the-art error reduction. The same model reduces pose kinematic velocity errors and acceleration errors.\\

The contributions of our work are as follows:

$\bullet$ A deeper CNN channel depth to reduce spatial and temporal positional error in the learned network by regressing skeletal joint keypoint data.

$\bullet$ Joint constraints as losses to improve positional and motion error reductions without increasing network complexity.

$\bullet$ Generalized improvements over action wise poses on the well known Human 3.6M benchmark data-set with a single model.
\\\\
Our model and code are available at https://vnmr.github.com/.

\section{Related}

There is a large corpus of literature on 3D human pose estimation. We briefly describe related work relevant to our method and results.

\paragraph {Features}

\cite{krizhevsky2012imagenet} establishes the importance of prior data in convolutional neural networks (CNN). The networks capacity as a feature detector is controlled by the depth and breadth of the architecture. However \cite{krizhevsky2012imagenet} does not address temporal feature detection across image frames. \cite{pavllo20193d} notes the importance of convolutional architectures in allowing precise control over temporal image frames. Dilated convolutions are introduced to model long term dependencies across the temporal receptive frame. When deciding a depth for the temporal convolutuonal neural network, \cite{pavllo20193d} study found that a depth of 1024 yielded the optimal balance of over and underfitting the detection of temporal features. Their approach learns over adjacent frames to improve temporal resolution. The question remains whether spatial and temporal accuracy of 3D pose estimators can be further improved by altering the CNN architecture with prior knowledge.

\paragraph{Spatial Poses}

A large body of human pose detection methods use feature detectors to accurately estimate 3D poses in an image.  \cite{sminchisescu20083d} \cite{ramakrishna2012reconstructing} \cite{ionescu2014iterated} \cite{ionescu2014iterated} \cite{akhter2015pose} \cite{jiang2018data} use joint limit characterization to learn pose dependent models of joint limits to form a prior.  
End to end methods use CNNs as a basis for pose estimation.~\cite{li20143d}~\cite{tekin2016direct}~\cite{tekin2016direct}~\cite{martinez2017simple} directly estimating 3D poses from image data and 2D joint keypoints can be lifted to 3D representations \cite{jiang20103d}~\cite{martinez2017simple}~\cite{pavlakos2017coarse}~\cite{tekin2017learning}~\cite{chen20173d}~\cite{rayat2017exploiting}~\cite{pavlakos2018ordinal} 
Coarse to fine discretization can further improve accuracy. \cite{pavlakos2017coarse}~\cite{sun2019deep} use a series of high-to-low resolution networks to enhance spatial precision.  
\cite{chen2018cascaded} introduces a Cascaded Pyramid Network (CPN) that includes a two stage GlobalNet and RefineNet that resolves difficult keypoints such as occluded points with a second network.  
Other end to end networks \cite{zhao2019semantic} introduce Semantic Graph Convolutional Networks ~\cite{chen2019weakly}, and ~\cite{kocabas2019self} multi-view geometry architectures to improve 3D pose detection.  
These methods improve spatial pose estimation and enable extraction of robust 2D skeletons, however they do not address temporal accuracy explicitly.

\paragraph{Temporal Poses}

Recent work \cite{kanazawa2019learning} predicts the 3D mesh of the human body, hallucinating past and future 3D dynamics from a  single image to estimate 3D poses. \cite{pavllo20193d} uses spatial refinement of a cascading pyramid network \cite{chen2018cascaded} with the precise control over a temporal receptive field of dilated convolutions. \cite{pavllo20193d} notes dilated convolutions ~\cite{holschneider1990real} success in a number of temporal domains \cite{yu2015multi} \cite{zhou2016sparseness} \cite{kalchbrenner2016neural} by preserving long term dependencies and maintaining efficiency. In recent work, \cite{pavllo20193d}, \cite{dabral2018learning} use joint angles, body symmetry, bone lengths and ratios constraints as losses. A promising joint angle representation \cite{pavllo2019modeling} with quaternions and forward kinematics generates qualitatively realistic body motion.  
Prior aware approaches shows promising results, and we endevour to further this by simplifying the required constraints to ensure generalization across position and temporal action wise poses.

\paragraph{Ours}Our work aims to improve spatial pose detection, temporal accuracy, and action pose generalization in a \textit{single} temporal convolutional model.  We add joint aware priors and update the filter depth \cite{krizhevsky2012imagenet} as contributions that enhance spatial \cite{chen2018cascaded} and temporal based pose estimation \cite{pavllo20193d}.  We build upon \cite{pavllo20193d} by updating depth filters and joint aware prior constraints as losses. This new architecture learns a refined positional and temporal model resulting in improved error generalization.  Since our method demonstrates contributions that build upon, but are loosely coupled to previous convolutional architectures \cite{pavllo20193d}, these contributions can transfer to future convolutional architectures.

\section{Architecture}

\begin{figure}
  \includegraphics[width=\linewidth]{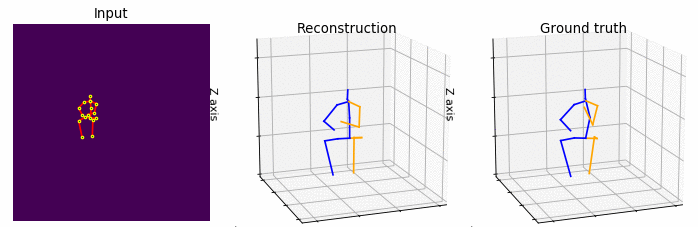}
  \caption{Reconstructed 3D Skeleton}
  \label{fig:skeleton}
\end{figure}

Our architecture builds upon the convolutional architecture introduced in \cite{pavllo20193d}. Our contribution updates the channel number in this network from the baseline architecture of 1024. Recall from \cite{krizhevsky2012imagenet} the depth of filter can significantly impact the feature recognition capacity of the network. Emperically we find that a channel depth of 2048 combined with the introduced loss functions yields optimal reduction in mean squared error in eq. (~\ref{eq:mse} ) between predicted $n_p$ values and ground truth target$n_t$ values. Once the 2D to 3D network identifies joint keypoints, a 3D skeleton is reconstructed \ref{fig:skeleton} and the following joint constraints as losses are applied. Positional constraints include joint angles between limbs, symmetry of limbs derived from joints on the left and right sides of the skeleton, and fixed distances between particular joints such as the distance between hip and knee joints on each side of the skeleton. The best performing emperically tested kinematic constraints are added to the position constraints including joint linear velocities, accelerations, and joint angular accelerations. 

\begin{table}
\centering
\caption{\label{tab:table-name}Loss Types}
\begin{tabular}{|l|l|}
\hline
\textbf{Loss} &
\textbf{Symbol} \\

\hline
        Angle Between Limbs &
        $\theta$
        \\
\hline
        Joint to Joint Fixed Distance &
        $d$
        \\
\hline
        Left vs Right Limb Symmetry&
        $s$
        \\
\hline
        Limb Range Of Motion &
        $r$
        \\
\hline
        Joint Linear Velocity &
        $\dot x$
        \\
\hline
        Joint Linear Acceleration &
        $\ddot x$
        \\
\hline
        Joint Angular Acceleration &
        $\ddot \theta$
        \\
\hline
\end{tabular}


\end{table}

\begin{equation} \label{eq:mse}
L = \sum_{n}^{N}\\
(n_p - n_t)^2|
_{n\in \{\theta, d, s, r, \dot x , \ddot x, \ddot \theta\}}
  \centering
\end{equation}

\section{Experiment}

Our network is trained on the Human 3.6M dataset \cite{ionescu2014iterated}. The dataset consists of 3.6 million 3D human poses and corressponding images. 11 human subjects perform 15 actions. Motion capture equipment is used to record 2D joint locations, 3D ground truth data, and 4 syncronized cameras capture video at 50 Hz (50 frames per second).

\begin{table}
\caption{Testbed}
\begin{center}
\small
\begin{tabular}{|l|l|l|}\hline
\textbf{Dataset} & Human 3.6M \\ \hline
\textbf{Training time} & $\approx$1hr/epoch\\ \hline
\textbf{Epochs} & 85 \\ \hline
\textbf{Processor} & Intel i9 8 core \\ \hline
\textbf{Accelerator} & Nvidia GeForce 2080ti \\ \hline
\end{tabular}
\end{center}
\label{table:testbed}
\end{table}

In keeping with previous reported Human 3.6M \cite{pavlakos2017coarse} \cite{tekin2017learning} \cite{martinez2017simple} \cite{sun2017compositional} \cite{fang2018learning} \cite{pavlakos2018ordinal} \cite{yang20183d} \cite{luvizon20182d} \cite{pavllo20193d} results, we continue to adopt a 17-joint skeleton, train on the same five subjects (S1, S5, S6, S7, S8) and test on two subjects (S9 and S11).

The standard Human 3.6M test protocols are evaluated as follows. \textit{Protocol 1} is the mean per-joint position error (MPJPE) in millimeters calculated as the Euclidean distance between predicted joint positions and ground-truth joint positions and follows \cite{li2015maximum} \cite{tekin2016direct} \cite{zhou2016sparseness} \cite{martinez2017simple} \cite{pavlakos2017coarse} \cite{pavllo20193d}. \textit{Protocol 2} is the error after alignment with the ground truth in translation, rotation, and scale (P-MPJPE) \cite{martinez2017simple} \cite{sun2017compositional} \cite{fang2018learning} \cite{pavlakos2018ordinal} \cite{yang20183d} \cite{rayat2017exploiting}. \textit{Protocol 3} is the error after aligning predicted poses with the ground-truth in scale only (N-MPJPE).  \cite{pavllo20193d} introduced temporally based motion metrics for velocity (MPJVE) which is the first derivative of the MPJPE 3D pose error.

We report our training results for the same outlined metrics using a codebase that reflects our new archcitecture. We also introduce the second derivative of the MPJPE error as the joint acceleration error (MPJAE). We train a single model for all actions for 85 epochs. In order to compare our results with the baseline architecture in \cite{pavllo20193d} \textit{no fine tuning} is performed. Hyperparamaters values are carried over from \cite{pavllo20193d} with the expection of a deeper channel from 1024 in \cite{pavllo20193d} to 2048 in ours as outlined in Table ~\ref{table:hyperparameters}.

In order to validate our architecture changes against the baseline \cite{pavllo20193d} we first downloaded their publicly available code base and reproduced their training results. We updated the architecture from 1024 channel depth to 2048, and instrumented the joint constraints as losses, trained the new network against the same dataset and compared our results. The testbed is outlined in Table ~\ref{table:testbed}, where each epoch takes approximately 1 hour to train.

\begin{table}
\caption{Hyperparameters}
\begin{center}
\small
\begin{tabular}{|l|l|l|l|}
\hline

\textbf{Parameter} &
\textbf{Value}\\
\hline
        Learning Rate&
        0.95
        \\
\hline
        Learning Rate Decay&
        0.95
        \\
\hline
        Drop Out &
        0.25
        \\
\hline
        Channel &
        2048
        \\
\hline
        Batch Size &
        1024
        \\
\hline
        CPN Architecture&
        3,3,3,3,3 243 frames
        \\
\hline
\end{tabular}
\end{center}
\label{table:hyperparameters}
\end{table}

\section{Results}

Results for our model trained on the Human3.6M dataset are as follows:

Baseline training results for channel size 2048 do not significantly differ with results reported in \cite{pavllo20193d}.
Training results for our joint aware model with a channel size of 2048 show a significant error reduction during training over the baseline. This indicates that updating the channel size to 2048 with the addition of our loss functions to the baseline architecture improves our test validation error rates.

\begin{table}
\caption{Actions}
\begin{center}
\small
\begin{tabular}{|l|l|}
\hline
Dir. & Directions \\ \hline 
Dis. & Discussion \\ \hline 
Eat  & Eating \\ \hline 
Grt & Greeting \\ \hline 
Phn & Phoning \\ \hline 
Pht & Photo \\ \hline 
Pos & Posing \\ \hline
Pur & Purchasing \\ \hline
Sit & Sitting \\ \hline
SitD & SittingDown \\ \hline
Smk & Smoking \\ \hline
Wat & Waiting \\ \hline
WD & Walk Dog \\ \hline
Wlk & Walking \\ \hline
WT & Walk Together \\ \hline
\end{tabular}
\end{center}
\label{fig:actions}
\end{table}

Test time results in Figure ~\ref{table:mpjpe}, Figure ~\ref{table:p-mpjpe}, Figure ~\ref{table:n-mpjpe} show our model outperforms all previous approaches in reducing mean per joint error for all three protocols. It also shows broad error reduction on a per action basis. This indicates good generalization of the model across broad sets of actions.  Test time results in Figure ~\ref{table:velocity} and Figure ~\ref{table:acceleration} show our average motion error reduction outperforms motion error rates from the reconstrcuted codebase of \cite{pavllo20193d} by 7\% for MPJVE and 13\% for MPJAE.

\begin{table}
\caption{Mean Per Joint Error Summary}
\begin{center}


\small
\begin{tabular}{|l|l|l|l|l|l|}\hline
& Protocol \#1 & Protocol \#2 & Protocol \#3 \\ \hline
Pavllo'19 (CVPR) & 46.8 & 36.5 & 44.9 \\ \hline
Chen'19 (CVPR) & 46.3 & 41.6 & 50.3 \\ \hline
\textbf{Ours} & \textbf{45.9} & \textbf{35.9} & \textbf{44.2}\\ \hline
\end{tabular}
\end{center}
\label{table::mpjpe-summary}
\end{table}

\begin{table}
\caption{Motion Summary}
\begin{center}
\small
\begin{tabular}{|l|l|l|}\hline
& Velocity & Acceleration\\ \hline
Pavllo'19 (CVPR) & 2.83 & 2.44\\ \hline
\textbf{Ours} & \textbf{2.63} & \textbf{2.12}\\ \hline
\end{tabular}
\end{center}
\label{table:motion-summary}
\end{table}

\begin{table}
\caption{Protocol 1 MPJPE}
\begin{center}
\tiny
\begin{tabular}{|l|l|l|l|l|l|l|l|l|l|l|l|l|l|l|l|l| }\hline

 & Dir.& Dis.& Eat & Grt & Phn & Pht& Pos& Pur & Sit& SitD & Smk & Wat & WD. & Wlk & WT & Avg \\ \hline




Fang'18 (AAAI)& 50.1 & 54.3 & 57.0 & 57.1 & 66.6 & 73.3 & 53.4 & 55.7 & 72.8 & 88.6 & 60.3 & 57.7 & 62.7 & 47.5 & 50.6 & 60.4 \\ \hline

Pavlakos'18 (CVPR) & 48.5 & 54.4 & 54.4 & 52.0 & 59.4 & 65.3 & 49.9 & 52.9 & 65.8 & 71.1 & 56.6 & 52.9 & 60.9 & 44.7 & 47.8 & 56.2 \\ \hline

Yang'18 (CVPR) & 51.5 & 58.9 & 50.4 & 57.0 & 62.1 & 65.4 & 49.8 & 52.7 & 69.2 & 85.2 & 57.4 & 58.4 & 43.6 & 60.1 & 47.7 & 58.6 \\ \hline

Luvizon '18 (CVPR)& 49.2 & 51.6 & 47.6 & 50.5 & 51.8 & 60.3 & 48.5 & 51.7 & 61.5 & 70.9 & 53.7 & 48.9 & 57.9 & 44.4 & 48.9 & 53.2 \\ \hline

Hos. '18 (ECCV)& 48.4 & 50.7 & 57.2 & 55.2 & 63.1 & 72.6 & 53.0 & 51.7 & 66.1 & 80.9 & 59.0 & 57.3 & 62.4 & 46.6 & 49.6 & 58.3 \\ \hline

Lee'18 (ECCV)& \textbf{40.2}  & 49.2  & 47.8  & 52.6  & 50.1  & 75.0  & 50.2  & \underline{43.0}  & \underline{55.8}  & 73.9 & 54.1 & 55.6 & 58.2 & 43.3 & 43.3 & 52.8 \\ \hline

Zhao '19 (CVPR) & 47.3 & 60.7 & 51.4 & 60.5 & 61.1 & 49.9 & 47.3 & 68.1 & 86.2 & 55.0 & 67.8 & 61.0 & 42.1 & 60.6 & 45.3 & 57.6 \\ \hline
Habibie '19 (CVPR) & 46.1 &  51.3 & 46.8 & 51.0 & 55.9 & \underline{43.9} & 48.8 & 65.8 & 81.6 & \underline{52.2} & 59.7 &51.1 & 40.8 & 54.8 & 45.2 & 53.4  \\ \hline
Li '19 (CVPR)  & 43.8 & 48.6 & 49.1 & 49.8 & 57.6 &  61.5 &  45.9 &  48.3 &  62.0 &  73.4 &  54.8 & 50.6 & 56.0 & 43.4 & 45.5 & 52.7  \\ \hline
Pavllo '19 (CVPR) & 45.1 & 46.7 & \textbf{42.0} & \underline{45.6} & 48.1 & 55.1 & 44.5 & 44.3 & 57.2 & 65.8 & \underline{47.1} & 44.0 & \underline{49.0} & \underline{32.6} & \underline{33.9} & 46.8\\ \hline
Chen'19 (CVPR) & \underline{41.1} & \textbf{44.2} & 44.9 & 45.9 & \textbf{46.5} & \textbf{39.3} & \textbf{41.6} & 54.8 & 73.2 & \textbf{46.2} & 48.7 & \textbf{42.1} & \textbf{35.8} & 46.6 & 38.5 & \underline{46.3} \\ \hline

%
%
\textbf{Ours} &

43.7 &

\underline{45.7} &

\underline{42.4} &

\textbf{44.5} &

\underline{47.8} &

54.9 &

\underline{42.7} &

\textbf{42.9} &

\textbf{55.7} &

64.9 &

\textbf{46.2} &

\underline{43.2} &

\underline{48.0} &

\textbf{32.2} &

\textbf{33.3} &

\textbf{45.9}

\\ \hline

\end{tabular}
\end{center}
\label{table:mpjpe}
\end{table}

\begin{table}
\caption{Protocol 2 P-MPJPE}
\begin{center}

\tiny
\begin{tabular}{|l|l|l|l|l|l|l|l|l|l|l|l|l|l|l|l|l| }\hline

 & Dir.& Dis.& Eat & Grt & Phn & Pht& Pos& Pur & Sit& StD & Smk & Wat & WD. & Wak & WT & Avg \\ \hline

Fang'18 (AAAI)& 38.2 & 41.7 & 43.7 & 44.9 & 48.5 & 55.3 & 40.2 & 38.2 & 54.5 & 64.4 & 47.2 & 44.3 & 47.3 & 36.7 & 41.7 & 45.7 \\ \hline

Pavlakos'18 (CVPR) & 34.7 & 39.8 & 41.8 & 38.6 & 42.5 & 47.5 & 38.0 & 36.6 & 50.7 & 56.8 & 42.6 & 39.6 & 43.9 & 32.1 & 36.5 & 41.8 \\ \hline

Yang'18 (CVPR) & \textbf{26.9} & \textbf{30.9} & 36.3 & 39.9 & 43.9 & 47.4 & \textbf{28.8} & \textbf{29.4} & \textbf{36.9} & 58.4 & 41.5 & \textbf{30.5} & \textbf{29.5} & 42.5 & 32.2 & 37.7 \\ \hline

Hos .???'18 (ECCV)& 35.7 & 39.3 & 44.6 & 43.0 & 47.2 & 54.0 & 38.3 & 37.5 & 51.6 & 61.3 & 46.5 & 41.4 & 47.3 & 34.2 & 39.4 & 44.1 \\ \hline

Li'19 (CVPR)  & 35.5 & 39.8 & 41.3 & 42.3 & 46.0 & 48.9 & 36.9 & 37.3 & 51.0 & 60.6 & 44.9 & 40.2 &  44.1 & 33.1 & 36.9 & 42.6  \\ \hline
Chen'19 (CVPR)& 36.9 & 39.3 & 40.5 & 41.2 & 42.0 & \textbf{34.9} & 38.0 & 51.2 & 67.5 & \textbf{42.1} & 42.5 & 37.5 & \underline{30.6} & 40.2 & 34.2 & 41.6 \\ \hline

Pavllo'19 (CVPR) & 34.1 & 36.1 & \textbf{33.9} & \underline{37.2} & \textbf{36.4} & \underline{42.2} & 34.4 & 33.5 & 45.0 & 52.5 & 37.4 & 33.8 & 37.8 & \underline{25.6} & \underline{27.3} & \underline{36.5}\\ \hline


\textbf{Ours} &

\underline{33.2} &

\underline{35.8} &

\underline{33.8} &

\textbf{36.3} &

\underline{36.6} &

42.0 &

\underline{32.9} &

\underline{32.4} &

\underline{44.2} &

\underline{51.7} &

\textbf{37.0} &

\underline{33.0} &

37.1 &

\textbf{25.1} &

\textbf{26.9} &

\textbf{35.9}
\\ \hline

\end{tabular}

\end{center}
\label{table:p-mpjpe}
\end{table}

%
%

\begin{figure*}
\begin{center}

\tiny
\begin{tabular}{|l|l|l|l|l|l|l|l|l|l|l|l|l|l|l|l|l| }\hline
 & Dir.& Dis.& Eat & Grt & Phn & Pht& Pos& Pur & Sit& StD & Smk & Wat & WD. & Wak & WT & Avg \\ \hline

Chen'19 (CVPR) & 45.9 & 48.0 & 48.6 & 50.8 & 48.9 & 45.1 & 46.1 & 57.4 & 77.3 & 49.4 &  54.2 & 47.2 & 39.9 & 49.9 & 42.9 & 50.3 \\ \hline


Pavllo'19 (CVPR) &
42.1 &
44.4 &
41.4 &
44.0 &
46.2 &
53.7 &
42.4 &
42.0 &
\textbf{54.8} &
63.6 &
45.2 &
41.9 &
46.3 &
31.2 &
31.8 &
44.7
\\ \hline

%
%

\textbf{Ours} &

\textbf{41.5} &

\textbf{44.2} &

\textbf{41.2} &

\textbf{43.4} &

\textbf{46.0} &

\textbf{53.1} &

\textbf{41.6} &

\textbf{41.2} &

\underline{53.7} &

\textbf{62.4} &

\textbf{44.5} &

\textbf{41.8} &

\textbf{45.8} &

\textbf{31.0} &

\textbf{31.6} &

\textbf{44.2}

\\ \hline

\end{tabular}

\end{center}
\label{table:n-mpjpe}
\caption{Protocol 3 N-MPJPE}
\end{figure*}

\begin{table}
\caption{MPJVE - Velocity}
\begin{center}

\scriptsize

\begin{tabular}{|l|l|l|l|l|l|l|l|l|l|l|l|l|l|l|l|l| }\hline
 & Dir.& Dis.& Eat & Grt & Phn & Pht& Pos& Pur & Sit& StD & Smk & Wat & WD. & Wak & WT & Avg \\ \hline

Pavllo'19 (CVPR) & 3.0 & 3.1 & 2.2 & 3.4 & 2.3 & 2.7 & 2.7 & 3.1 & 2.1 & 2.9 & 2.3 & 2.4 & 3.7 & 3.1 & 2.8 & 2.8 \\ \hline

%
%
\textbf{Ours}&
\textbf{2.8} &

\textbf{2.9} &

\textbf{2.1} &

\textbf{3.2} &

\textbf{2.2} &

\textbf{2.6} &

\textbf{2.5} &

\textbf{2.9} &

\textbf{1.9} &

\textbf{2.7} &

\textbf{2.2} &

\textbf{2.3} &

\textbf{3.5} &

\textbf{3.0} &

\textbf{2.7} &

\textbf{2.63}
\\ \hline

\end{tabular}

\end{center}
\label{table:velocity}

\end{table}

\begin{table}
\caption{MPJAE - Acceleration}
\begin{center}

\scriptsize
\begin{tabular}{|l|l|l|l|l|l|l|l|l|l|l|l|l|l|l|l|l| }\hline
 & Dir.& Dis.& Eat & Grt & Phn & Pht& Pos& Pur & Sit& StD & Smk & Wat & WD. & Wak & WT & Avg \\ \hline

Pavllo'19 (CVPR)& 2.3 & 2.6 & 1.8 & 2.7 & 2.0 & 2.3 & 2.1 & 2.5 & 2.1 & 2.1 & 2.3 & 2.1 & 2.1 & 2.8 & 2.6 & 2.44 \\ \hline

%
%
\textbf{Ours}&
\textbf{2.1} &

\textbf{2.4} &

\textbf{1.7} &

\textbf{2.4} &

\textbf{1.9} &

\textbf{2.0} &

\textbf{1.9} &

\textbf{2.3} &

\textbf{1.9} &

\textbf{2.5} &

\textbf{1.9} &

\textbf{1.9} &

\textbf{2.6} &

\textbf{2.4} &

\textbf{2.2} &

\textbf{2.12}
\\ \hline
\end{tabular}
\end{center}
\label{table:acceleration}
\end{table}

\begin{figure}
  \centering
\begin{tikzpicture}
\begin{axis}[
    xlabel={Epoch},
    ylabel={Protocol \#1 Error},
    legend pos=north east,
    xmajorgrids=true,
    ymajorgrids=true,
    grid style=dashed,
    ]
    \addplot[
      color=black,
      mark=triangle,
    ]
    coordinates {
      (5,50.508422)
      (10,48.633593)
      (15,48.915212)
      (20,48.654959)
      (25,48.122122)
      (30,48.221232)
      (35,48.009611)
      (40,47.883297)
      (45,47.832548)
      (50,47.694754)
      (55,47.851320)
      (60,47.693382)
      (65,47.801333)
      (70,47.748820)
      (75,47.791186)
      (80,47.672664)
    };
    \addplot[
      color=black,
      mark=square,
    ]
    coordinates {
      (5,49.567341)
      (10, 48.0)
      (15,48.824429)
      (20,47.754615)
      (25,47.570226)
      (30,47.555887)
      (35,47.328202)
      (40,47.165046)
      (45,47.165962)
      (50,47.178484)
      (55,47.115763)
      (60,47.098911)
      (65,47.265339)
      (70,47.120916)
      (75,47.059782)
      (80,46.994238)
      (83,46.893375)
    };
    \legend{Pavllo'19 CVPR,
        Ours}
\end{axis}
\end{tikzpicture}
\caption{2048 Channels}
\end{figure}
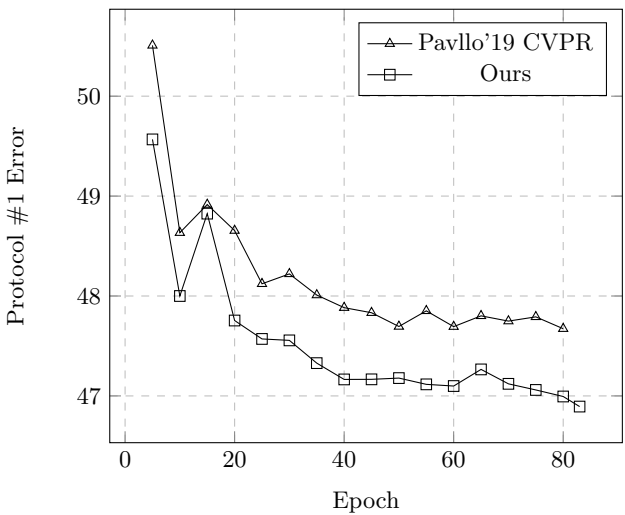

\section{Conclusion}

We introduced a joint aware fully convolutional model for 3D human pose estimation in video. An architecture with joint constraints as losses demonstrates improved mean results over actions on the Human3.6M dataset for all protocols and motion metrics resulting in new state-of-art-results. Our contribution to increase positional accuracy and motion smoothness in video can be integrated with future end to end networks without increasing network complexity.


%


\par\vfill\par

\clearpage
%
%

\bibliographystyle{splncs04}
\bibliography{egbib}
\end{document}